\documentclass[nohyperref]{article}
\usepackage{microtype}
\usepackage{graphicx}
\usepackage{subfigure}
\usepackage{booktabs} 
\usepackage{xcolor}   
\usepackage[utf8]{inputenc} 
\usepackage[T1]{fontenc}    
\definecolor{mydarkblue}{rgb}{0,0.08,0.45}
\usepackage[pagebackref,
    colorlinks=true,
    linkcolor=mydarkblue,
    citecolor=mydarkblue,
    filecolor=mydarkblue,
    urlcolor=mydarkblue]{hyperref}

\usepackage{url}            
\usepackage{booktabs}       
\usepackage{amsfonts}       
\usepackage{nicefrac}       
\usepackage{microtype}      

\usepackage{sidecap}
\usepackage{float}
\usepackage{graphicx}

\usepackage{mathrsfs}

\usepackage{amsmath}
\usepackage{amssymb}
\usepackage{amsthm}
\usepackage{mathtools}
\usepackage[mathscr]{eucal}%
\usepackage{bm}
\usepackage{bbm}
\usepackage{relsize}
\usepackage{graphics}
\usepackage{wrapfig}
\usepackage{multirow}
\usepackage{enumitem}
\usepackage{tablefootnote}

\usepackage{array}
\usepackage{color, colortbl}
\definecolor{Gray}{gray}{0.9}
\usepackage{algorithm}
\usepackage{algorithmic}
\usepackage{pgf}
\usepackage{xinttools}
\usepackage{tikz}
\usetikzlibrary{arrows.meta}
\usepackage{textcomp}
\usetikzlibrary{calc,shapes,arrows,positioning,automata,trees}
\usepackage{placeins} 
\usepackage{tabularx}
\usepackage{graphicx}

\usepackage[capitalize,noabbrev]{cleveref}
\usepackage{cancel}   
\def\ie{\emph{i.e.}}

\def\etal{{\em et al.}}

\usepackage{bigdelim}
\usepackage{colortbl} 
\definecolor{mColor1}{rgb}{0.95,0.95,0.95}

\usepackage{microtype}
\usepackage{graphicx}
\usepackage{booktabs} 

\graphicspath{{figure/}}

\usepackage{soul}

\usepackage[accepted]{icml2023}

\usepackage[textsize=tiny]{todonotes}

\icmltitlerunning{MetaModulation: Learning Variational Feature Hierarchies for Few-Shot Learning with Fewer Tasks}

\begin{document}

\twocolumn[
\icmltitle{MetaModulation: Learning Variational Feature Hierarchies \\for Few-Shot Learning with Fewer Tasks}
%
 
\icmlsetsymbol{equal}{*}

\begin{icmlauthorlist}
\icmlauthor{Wenfang Sun}{equal,cas,ustc}
\icmlauthor{Yingjun Du}{equal,uva}
\icmlauthor{Xiantong Zhen}{uih}
\icmlauthor{Fan Wang}{cas}
\icmlauthor{Ling Wang}{cas}
\icmlauthor{Cees G.M. Snoek}{uva}
\end{icmlauthorlist}

\icmlaffiliation{cas}{ Hefei Institutes of Physical Science, Chinese Academy of Sciences, Hefei 230031, China/P. R. China.}
\icmlaffiliation{ustc}{University of Science and Technology of China, Hefei 230026, China/P. R. 
China.}
\icmlaffiliation{uva}{University of Amsterdam, Amsterdam, the Netherlands.}
\icmlaffiliation{uih}{United Imaging Healthcare,  Beijing, China.}

\icmlcorrespondingauthor{Wenfang Sun}{swf@mail.ustc.edu.cn}
\icmlcorrespondingauthor{Yingjun Du}{y.du@uva.nl}


\icmlkeywords{Machine Learning, ICML}

\vskip 0.3in
]




\printAffiliationsAndNotice{\icmlEqualContribution} 
\begin{abstract}
Meta-learning algorithms are able to learn a new task using previously learned knowledge, but they often require a large number of meta-training tasks which may not be readily available. To address this issue, we propose a method for few-shot learning with fewer tasks,  which we call MetaModulation. 
The key idea is to use a neural network to increase the density of the meta-training tasks by modulating batch normalization parameters during meta-training. 
Additionally, we modify parameters at various network levels, rather than just a single layer, to increase task diversity. To account for the uncertainty caused by the limited training tasks, we propose a variational MetaModulation where the modulation parameters are treated as latent variables. 
We also introduce learning variational feature hierarchies by the variational MetaModulation, which modulates features at all layers and can consider task uncertainty and generate more diverse tasks.
The ablation studies illustrate the advantages of utilizing a learnable task modulation at different levels and demonstrate the benefit of incorporating probabilistic variants in few-task meta-learning. Our MetaModulation and its variational variants consistently outperform state-of-the-art alternatives on four few-task meta-learning benchmarks. 
\end{abstract}

\section{Introduction}
\label{sec:intro}
Learning to learn or \textit{meta-learning}~\cite{schmidhuber1987evolutionary, thrun1998learning}, offers a powerful tool for few-shot learning~\cite{andrychowicz2016learning, ravi2016optimization, finn2017model}. The crux for few-shot meta-learning is to accrue prior meta-knowledge from a set of meta-training tasks, which enables fast adaptation to a new task with limited data.  
Despite remarkable achievements of existing meta-learning algorithms for few-shot learning~\cite{finn2017model, snell2017prototypical, Liu_2022_CVPR, Hu_2022_CVPR, He_2022_CVPR} 
these works depend on a large number of meta-training tasks during training. However, an extensive collection of meta-training tasks is unlikely to be available for many real-world applications. For example, in medical image diagnosis, a shortage of data samples and tasks arises due to the need for specialist labeling by physicians and patient privacy concerns. Additionally, rare disease types \cite{wang2017chestx} present challenges for few-shot learning. In this paper, we focus on few-task meta-learning, where the number of available tasks at training time is limited. 

\par
To tackle the few-task meta-learning problem, a variety of task augmentation~\cite{ni2021data, yao2021improving} and task interpolation~\cite{lee2022set, yao2021meta} methods have been proposed. The key idea of task augmentation~\cite{ni2021data, yao2021improving} is to increase the number of tasks from the support set and query set during meta-training. The weakness of these approaches is that they are only able to capture the global task distribution within the distribution of the provided tasks. Task interpolation~\cite{lee2022set, yao2021meta} generates a new task by interpolating the support and query sets of different tasks by Mixup~\cite{verma2019manifold} or a neural set function~\cite{lee2019set}. Here, a key question is how to combine tasks and at what feature level. For example, the state-of-the-art MLTI by~\cite{yao2021meta} randomly selects the features of a single layer from two known tasks for a linear mixup but ignores all other feature layers for new task generation. It leads to a sub-optimal interpolated task diversity.  
To address this limitation, we propose a new task modulation strategy that captures the knowledge from one known task at different levels. 

\par 
One key aspect of task modulation is the ability to leverage the representation of a single task at different levels of abstraction. This allows the model to modulate representations of other tasks at varying levels of detail, depending on the specific needs of the new task. 
Conditional batch normalization~\cite{de2017modulating,dumoulin2016learned,perez2018film} has been successfully applied to visual question answering and other multi-modal applications. 
In conditional batch normalization,  the normalization parameters (i.e., the scale and shift parameters) are learned from a set of additional input conditions, which can be represented as a set of auxiliary variables or as a separate input branch to the network.
This allows the network to adapt to the specific task at hand and improve its performance. Inspired by these general-purpose conditional batch normalization methods, we make in this paper three contributions.

In this paper, we propose a method for few-shot learning with fewer tasks called MetaModulation. It contains three key contributions. First, a meta-training task is randomly selected as a base task, and additional task information is introduced as a condition.
We predict the scale and shift of the batch normalization for the base task from the conditional task. This allows the model to modulate the statistics of the conditional task on the base task for a more effective task representation. It is also worth noting that our modulation operates on each layer of the neural network, while previous methods~\cite{yao2021meta, lee2022set} only select a single layer for modulation. Thus, the model can generate more diverse tasks during meta-training, as it utilizes the statistical information of each level of the conditional task. 
As a second contribution, we introduce variational task modulation, which treats the conditional scale and shifts as latent variables inferred from the conditional task. The optimization is formulated as a variational inference problem, and new evidence lower bound is derived under the meta-learning framework. In doing so, the model obtains probabilistic conditional scale and shift values that are more informative and better represent the distribution of real tasks. 
As a third contribution, we propose hierarchical variational task modulation, which obtains the probabilistic conditional scale and shifts at each layer of the network. We cast the optimization as a hierarchical variational inference problem in the Bayesian framework; the inference parameters of the conditional scale and shift are jointly optimized in conjunction with the modulated task training.

\par
To verify our method, we conduct experiments on four few-task meta-learning benchmarks: miniImagenet-S, ISIC, DermNet-S, and Tabular Murris. We perform a series of ablation studies to investigate the benefits of using a learnable task modulation method at various levels of complexity. Our goal is to illustrate the advantages of increasing task diversity through such a method, as well as demonstrate the benefits of incorporating probabilistic variations in the few-task meta-learning framework. Our experiments show that MetaModulation consistently outperforms state-of-the-art few-task meta-learning methods on the four benchmarks.

\section{Preliminaries}
\paragraph{Problem statement.}
For the traditional few-shot meta-learning problem,  we deal with tasks $\mathcal{T}_i$, as sampled from a task distribution $p(\mathcal{T})$. We sample $N$-way $k$-shot tasks from the meta-training tasks, where $k$ is the number of labeled examples for each of the $N$ classes. Each  $t$-th task includes a support set $\mathcal{S}^{t}{=}\{(\mathbf{x}_i, \mathbf{y}_i)\}_{i=1}^{N\mathord\times k}$ and query set $\mathcal{Q}^{t}{=}\{(\tilde{\mathbf{x}}_i, \tilde{\mathbf{y}}_i)\}_{i=1}^m$ ($\mathcal{S}^{t}, \mathcal{Q}^{t} \subseteq \mathcal{X}$).   Given a learning model $f_{\phi}$, where $\phi$ denotes the model parameters, few-shot learning algorithms attempt to learn $\phi$ to minimize the loss on the query set $\mathcal{Q}_i$ for each of the sampled tasks using the data-label pairs from the corresponding support set $\mathcal{S}_i$. After that, during the testing stage, the trained model $f_{\phi}$ and the support set $\mathcal{S}_j$ for new tasks $\mathcal{T}_j$ perform inference and evaluate performance on the corresponding query set $\mathcal{Q}_j$.  
In this paper, we focus on \textit{few-task} meta-learning. In this setting, the main challenge is that the number of meta-training tasks $\mathcal{T}_i$ is limited, which causes the model to overfit easily.

\paragraph{Prototype-based meta-learning.}
We develop our method based on the prototypical network (ProtoNet) by ~\citet{snell2017prototypical}. Specifically, ProtoNet leverages a non-parametric classifier that assigns a query point to the class having the nearest prototype in the learned embedding space. The prototype $\mathbf{c}_k$ of an object class $c$ is obtained by: $\mathbf{c}_k {=} \frac{1}{K}\sum_k f_{\phi}(\mathbf{x}_{c,k})$, where $f_{\phi}(\mathbf{x}_{c,k})$ is the feature embedding of the sample $\mathbf{x}_{c,k}$, which is usually obtained by a convolutional neural network. For each query sample $\mathbf{x}^q$, the distribution over classes is calculated based on the softmax over distances to the prototypes of all classes in the embedding space:
\begin{equation}
\label{eqn:protonetProb}
\footnotesize
    p(\mathbf{y}_{n}^q = k|\mathbf{x}^q) = \frac{\exp(-d(f_{\phi}(\mathbf{x}^q),\mathbf{c}_k))}{\sum_{k'} \exp(-d(f_{\phi}(\mathbf{x}^q),\mathbf{c}_{k'}))},
\end{equation}
where $\mathbf{y}^q$ denotes a random one-hot vector, with $\mathbf{y}_c^q$ indicating its $n$-th element, and $d(\cdot,\cdot)$ is some (Euclidean) distance function. Due to its non-parametric nature, the ProtoNet enjoys high flexibility and efficiency, achieving considerable success in few-shot learning.

\paragraph{Conditional batch normalization.}
The aim of  Batch Normalization~\cite{ioffe2015batch} is to accelerate the training of deep networks by reducing internal covariate shifts. For a layer with $d$-dimensional input $x{=}({x}^{(1)}...{x}^{(d)})$ and activation $x^{(k)}$, batch normalization normalizes each scalar feature as follows:
\begin{equation}\label{eqn:Batch Normalization}
\footnotesize
\mathbf{y}^{(k)}={\gamma}^{(k)}\frac{{x}^{(k)}-\mathbb{E}[{x}^{(k)}]}{\sqrt{\mathrm{Var}[{x}^{(k)}]+\epsilon}}+{\beta}^{(k)},
\end{equation}
where $\epsilon$ is a constant added to the variance for numerical stability.  ${{\gamma}^{(k)}}$ and ${{\beta}^{(k)}}$ are the scale and shift for batch normalization.
Conditional batch normalization (CBN)~\cite{de2017modulating} is a class-conditional variant of conventional batch normalization. The key idea of CBN is to predict the transformation parameters $\gamma$ and $\beta$ from a conditional embedding (\ie, a language embedding). CBN enables the language embedding to manipulate the entire vision feature map by scaling them up or down, negating them, or shutting them off completely. Specifically, CBN uses two feed-forward multi-layer perceptrons (MLPs) to predict these changes instead of predicting the original transformations, which benefits training stability:
\begin{equation}\label{eqn:MLP deltas}
\Delta\beta=\texttt{MLP}(\mathbf{e}_q)           \qquad       \Delta\gamma=\texttt{MLP}(\mathbf{e}_q),   
\end{equation}
where ${e}_q$ is an additional language embedding. So, given a feature map with $C$ channels, these MLPs 
 output a vector of size $C$. They then add these changes  to the $\beta$ and $\gamma$ parameters:
\begin{equation}\label{eqn:CBN}
\hat{\beta_c}=\beta_c+\Delta\beta_c     \qquad   \hat{\gamma_c}=\gamma_c+\Delta\gamma_c.
\end{equation}
Finally, the updated $\hat{\beta}$ and $\hat{\gamma}$ are used as transformation parameters for the batch normalization~(eq.~( \ref{eqn:Batch Normalization})) of vision features. Rather than using a language embedding for the conditioning, we randomly select one additional task as a condition to predict the scale and shift of the batch normalization for another task.

\paragraph{Meta-learning task interpolation.}
Several methods~\cite{yao2021meta, lee2022set} have been suggested as ways to increase the diversity of the tasks used for meta-training.
MLTI~\cite{yao2021meta} generates additional tasks by randomly sampling a pair of tasks and interpolating the corresponding features and labels using Manifold Mixup~\cite{verma2019manifold}. Specifically, given examples from class $n$ in task $\mathcal{T}_i$ and class $n'$ in task $\mathcal{T}_j$, the interpolated features are defined as:
\begin{equation}
\label{eq:nsl_mix_s}
\footnotesize
\hat{\mathbf{H}}^{s,l}_{n}=\lambda \mathbf{H}^{s,l}_{i;n}+(1-\lambda)\mathbf{H}^{s,l}_{j;n'},
\end{equation}
\begin{equation}
\footnotesize
\label{eq:nsl_mix_q}
\hat{\mathbf{H}}^{q,l}_{n}=\lambda \mathbf{H}^{q,l}_{i;n}+(1-\lambda)\mathbf{H}^{q,l}_{j;n'},
\end{equation}
where $l$ indicates the $l$-th layer ($0 \leq l \leq L$), and $\lambda \in \left[ 0, 1\right]$ is sampled from a Beta distribution $\mathrm{Beta}(\alpha, \beta)$. The interpolated support samples $\hat{\mathbf{H}}^{s,l}_{cn;n}$ and query samples $\hat{\mathbf{H}}^{q,l}_{cn;n}$ can be regarded as the new classes in the interpolated task. However, MLTI~\cite{yao2021meta} randomly selects only the features of a single layer from two known tasks to be mixed and ignores all the other feature layers. It leads to the interpolated task's diversity being limited and therefore does not increase the generalizability of the model.

\section{MetaModulation}
\label{sec:methods}

In this paper, we propose  MetaModulation for few-task meta-learning.  We first introduce meta task modulation in section~\ref{sec:cti}. To obtain more diverse meta-training tasks, we then propose variational task modulation in section~\ref{sec:cvti}, which introduces variational inference into the modulation. We also introduce  hierarchical meta variational modulation in section~\ref{sec:chvti}, which adds  variational modulation to each network layer to obtain a richer task distribution.

\subsection{Meta task modulation}
\label{sec:cti}
To address the single layer limitation in MLTI~\cite{yao2021meta}, we introduce meta task modulation for few-task meta-learning, which 
modulates the features of two different tasks at different layers.
%
%
We modulate all layers of samples from a meta-training task $\mathcal{T}_j$ by predicting the $\gamma$ and $\beta$ of the batch normalization from base task $\mathcal{T}_i$. Following CBN~\cite{de2017modulating}, we only predict the change $\Delta\beta_c$ and $\Delta\gamma_c$ on the original scalars from the task $\mathcal{T}_i$, which benefits training stability.

Specifically, to infer the conditional scale and shift $\Delta\beta_c$ and $\Delta\gamma_c$, we deploy two functions $f_{\beta}^l(\cdot)$ and  $f_{\gamma}^l(\cdot)$ that take the activations $\mathbf{H}^{l}_{i;n}$  from task $\mathcal{T}_i$ as input, and the output are $\Delta\beta^{l}_{i;n;c}$ and $\Delta\gamma^{l}_{i;n;c}$. The functions $f_{\beta}^l(\cdot)$ and  $f_{\gamma}^l(\cdot)$ are  parameterized by two feed-forward multi-layer perceptrons:
\begin{equation}\label{eqn:cti_deltas}
\Delta\beta^{s,l}_{i;n;c}= \texttt{MLP}(\mathbf{H}^{s,l}_{i;n})           \qquad       \Delta\gamma^{s,l}_{i;n;c}=\texttt{MLP}(\mathbf{H}^{s,l}_{i;n})   
\end{equation}
where $\Delta\gamma^{s,l}_{i;n;c}$ and $\Delta\gamma^{s,l}_{i;n;c}$ are the changes of the support set. We obtain   $\Delta\gamma^{q,l}_{i;n;c}$ and $\Delta\gamma^{q,l}_{i;n;c}$ of the query set by the same strategy. Note that the functions $f_{\beta}^l(\cdot)$ and  $f_{\gamma}^l(\cdot)$ are shared by different channels in  same layer and we learn $L$ pairs of those functions if we have $L$ convolutional layers. 

\begin{figure} [t]
\centering
\includegraphics[width=0.85\linewidth]{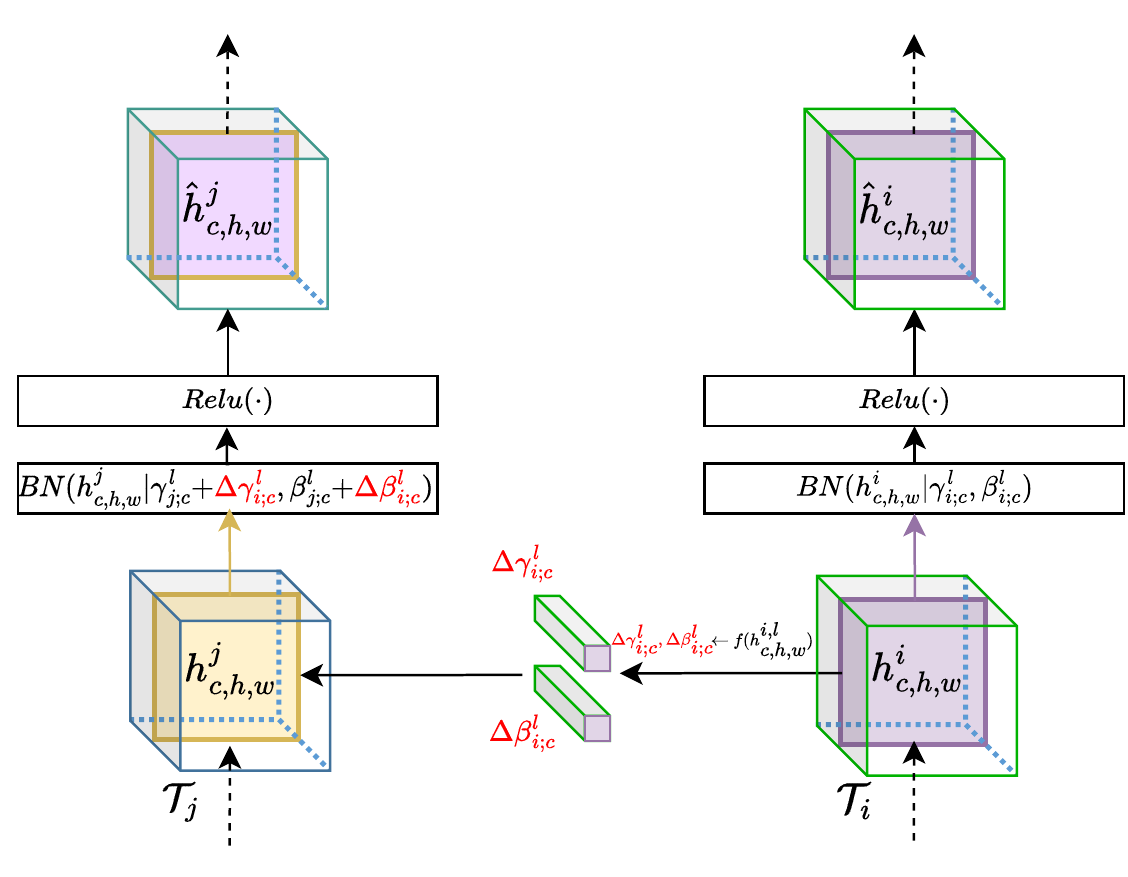}
\vspace{-4mm}
\caption{
\textbf{Meta task modulation.} Various combinations of the transformation parameters $\gamma$ and $\beta$ from task $\mathcal{T}_i$ can modulate the individual activation of task $\mathcal{T}_j$ at different layers, 
which can make the newly generated task more diverse.}
\vspace{-5mm}
	\label{fig:framework}
\end{figure}

Using the above functions, we generate the changes for the batch normalization scale and shift, then following eq.~(\ref{eqn:CBN}), we add these changes to the original $\beta^{l}_{j;n;c}$ and $\gamma^{l}_{j;n;c}$ from task $\mathcal{T}_j$:
\begin{equation}\label{eqn:cti_CBN}
\hat{\beta}^{s,l}_{j;n;c} =\beta^{s,l}_{j;n;c}  +\Delta\beta^{s,l}_{i;n;c}    \qquad \hat{\gamma}^{s,l}_{j;n;c} =\gamma^{s,l}_{j;n;c}  +\Delta\gamma^{s,l}_{i;n;c} 
\end{equation}
Once we obtain the modulated  scale $\hat{\gamma}^{l}_{i;n;c}$ and shift $\hat{\beta}^{l}_{i;n;c}$, we  compute the modulated features for the support and query set from task $\mathcal{T}_j$ based on eq. (\ref{eqn:Batch Normalization}):
\begin{equation}
\footnotesize
\label{eq:cti_mix_s}
\hat{\mathbf{H}}^{s,l}_{n}=  
\hat{\gamma}^{s,l}_{i;n;c} \frac{\mathbf{H}^{s,l}_{j;n} -\mathbb{E}[\mathbf{H}^{s,l}_{j;n}]}{\sqrt{\mathrm{Var}[\mathbf{H}^{s,l}_{j;n}]+\epsilon}}+ \hat{\beta}^{s,l}_{j;n;c},
\end{equation}

\begin{equation}
\footnotesize
\label{eq:cti_mix_Q}
\hat{\mathbf{H}}^{q,l}_{n}=  
\hat{\gamma}^{q,l}_{i;n;c} \frac{\mathbf{H}^{q,l}_{j;n} -\mathbb{E}[\mathbf{H}^{q,l}_{j;n}]}{\sqrt{\mathrm{Var}[\mathbf{H}^{q,l}_{j;n}]+\epsilon}}+ \hat{\beta}^{q,l}_{j;n;c},
\end{equation}
where $\mathbb{E}[\mathbf{H}^{l}_{i;n}]$ and $\mathrm{Var}[\mathbf{H}^{l}_{i;n}]$ are the mean and variance of samples features from $\mathcal{T}_j$.  We illustrate the meta task modulation process in Figure~\ref{fig:framework}. 

However, the deterministic conditional scale and shift are not sufficiently representative of modulated tasks. Moreover, uncertainty is inevitable due to the scarcity of data and tasks, which should also be encoded into the conditional scale and shift. In the next section, we derive a probabilistic latent variable model by modeling conditional scale and shift as distributions, which we learn by variational inference.

\subsection{Variational task modulation}
\label{sec:cvti}
In this section, we introduce variational task modulation using a latent variable model in which we treat the conditional scale 
$\Delta\beta^{s,l}_{i;n;c}$ and shift $\Delta\gamma^{s,l}_{i;n;c}$ as latent variables $\mathbf{z}$ inferred from one known task. 
We formulate the optimization of variational task modulation as a variational inference problem by deriving a new evidence lower bound (ELBO) under the meta-learning framework.

From a probabilistic perspective, the conditional latent scale and shift maximize the conditional predictive log-likelihood from two known tasks $\mathcal{T}_i, \mathcal{T}_j$.
\begin{equation}
\begin{aligned}
\label{eq:CVTI_like_3}
& \underset{p} \max \log p(\hat{\mathbf{y}} | \mathcal{T}_i, \mathcal{T}_j) \\
& = \underset{p} \max \log \int p(\hat{\mathbf{y}}|\hat{\mathbf{x}}^q, \hat{\mathbf{x}}^s) p(\hat{\mathbf{x}}^q, \hat{\mathbf{x}}^s | \mathcal{T}_i, \mathcal{T}_j)\mathrm{d}\hat{\mathbf{x}}^q \mathrm{d} \hat{\mathbf{x}}^s \\
& = \underset{p} \max \log\int p(\hat{\mathbf{y}} |\hat{\mathbf{x}}^q, \hat{\mathbf{x}}^s) p(\hat{\mathbf{x}}^q, \hat{\mathbf{x}}^s|\mathbf{z}, \mathcal{T}_j) p(\mathbf{z}|\mathcal{T}_i)\mathrm{d}\mathbf{z}\mathrm{d}\hat{\mathbf{x}}^q \mathrm{d} \hat{\mathbf{x}}^s
\end{aligned}
\end{equation}
where $\hat{\mathbf{x}}^s, \hat{\mathbf{x}}^q$ are the support sample and query sample of the modulated task $\mathcal{T}$. Since  \begin{small}$ p(\mathbf{z}, \hat{\mathbf{x}}^q, \hat{\mathbf{x}}^s | \mathcal{T}_i, \mathcal{T}_j) {=} p(\hat{\mathbf{x}}^q, \hat{\mathbf{x}}^s|\mathbf{z}, \mathcal{T}_j) p(\mathbf{z}|\mathcal{T}_i)$ \end{small} is generally intractable, we resort to a variational posterior $q(\mathbf{z}, \hat{\mathbf{x}}^q, \hat{\mathbf{x}}^s | \mathcal{T}_j)$ for its approximation.  We obtain the variational distribution by minimizing the Kullback-Leibler (KL) divergence:
\begin{equation}
 D_{\mathrm{KL}} [q(\mathbf{z}, \hat{\mathbf{x}}^q, \hat{\mathbf{x}}^s | \mathcal{T}_j) || p(\mathbf{z}, \hat{\mathbf{x}}^q, \hat{\mathbf{x}}^s | \mathcal{T}_i, \mathcal{T}_j). 
\label{eq:first_kl}
\end{equation}

\begin{figure} [t]
\centering
\includegraphics[width=0.9\linewidth]{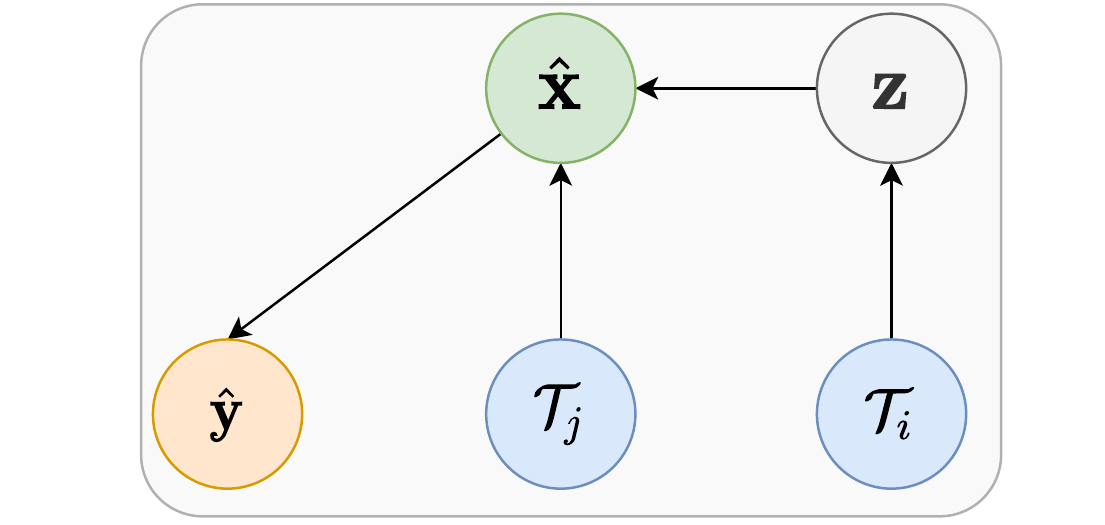}
\caption{\textbf{Variational task modulation.} $\hat{\mathbf{x}}$ and $\hat{\mathbf{y}}$ denote the sample and label of newly generated task $\hat{\mathcal{T}}$ and $\mathbf{z}$ represents the latent modulation parameters.} 
	\label{fig:vi_graph}
	\vspace{-6mm}
\end{figure}

By applying the Baye’s rule to the posterior \begin{small} $q(\mathbf{z}, \hat{\mathbf{x}}^q, \hat{\mathbf{x}}^s | \mathcal{T}_i)$ \end{small}, we derive the ELBO as:
\begin{equation}
\small
\label{eq:CVTI_elbo_1}
\begin{aligned}
 \log p(\hat{\mathbf{y}} | \mathcal{T}_i, \mathcal{T}_j) 
 \geq & \mathbb{E}_{q(\mathbf{z}, \hat{\mathbf{x}}^q, \hat{\mathbf{x}}^s)} \left[\log p(\hat{\mathbf{y}}| \hat{\mathbf{x}}^q, \hat{\mathbf{x}}^s)\right]  \\
& -  D_{\mathrm{KL}} \left[q(\mathbf{z}, \hat{\mathbf{x}}^q, \hat{\mathbf{x}}^s | \mathcal{T}_j) || p(\mathbf{z}, \hat{\mathbf{x}}^q, \hat{\mathbf{x}}^s | \mathcal{T}_i, \mathcal{T}_j)\right] \\
\end{aligned}
\end{equation}
The second term in the ELBO can also be simplified. Since 
\begin{equation}
\label{eq:CVTI_elbo_1_mid}
    \begin{aligned}
    & D_{\mathrm{KL}} \left[q(\mathbf{z}, \hat{\mathbf{x}}^q, \hat{\mathbf{x}}^s) | \mathcal{T}_i || p(\mathbf{z}, \hat{\mathbf{z}} | \mathcal{T}_i, \mathcal{T}_j)\right] \\ 
    & = \mathbb{E}_{q(\mathbf{z}, \hat{\mathbf{x}}^q, \hat{\mathbf{x}}^s)} \log \frac{q(\mathbf{z}, \hat{\mathbf{x} | \mathcal{T}_i})}{p(\mathbf{z}, \hat{\mathbf{x} | \mathcal{T}_i}, \mathcal{T}_j)}, 
    \end{aligned}
\end{equation}
and 
\begin{equation}
\label{eq:CVTI_elbo_1_mid_1}
    \begin{aligned} q(\mathbf{z}, \hat{\mathbf{x}}^q, \hat{\mathbf{x}}^s | \mathcal{T}_j) = p(\hat{\mathbf{x}}^q, \hat{\mathbf{x}}^s | \mathbf{z}, \mathcal{T}_j) q(\mathbf{z}),
        \end{aligned}
\end{equation}
we then combine eq.~(\ref{eq:CVTI_elbo_1_mid}), eq.~(\ref{eq:CVTI_elbo_1_mid_1}) and eq.~(\ref{eq:CVTI_like_3}), to obtain:
\begin{equation}
\label{eq:CVTI_elbo_mid_1}
    \begin{aligned}
    & \mathbb{E}_{q(\mathbf{z}, \hat{\mathbf{x}}^q, \hat{\mathbf{x}}^s)} \log \frac{q(\mathbf{z}, \hat{\mathbf{x}}^q, \hat{\mathbf{x}}^s | \mathcal{T}_j)}{p(\mathbf{z}, \hat{\mathbf{x}}^q, \hat{\mathbf{x}}^s | \mathcal{T}_i, \mathcal{T}_j)} \\
    & =  \mathbb{E}_{q(\mathbf{z}, \hat{\mathbf{x}}^q, \hat{\mathbf{x}}^s)} \log \frac{ {p(\hat{\mathbf{x}}^q, \hat{\mathbf{x}}^s | \mathbf{z}, \mathcal{T}_i)}q(\mathbf{z})}{{p(\hat{\mathbf{x}}^q, \hat{\mathbf{x}}^s | \mathbf{z}, \mathcal{T}_i)}p(\mathbf{z}|\mathcal{T}_i)} \\
    & =  \mathbb{E}_{q(\mathbf{z})} \log \frac{q(\mathbf{z})}{p(\mathbf{z}|\mathcal{T}_i)} \\
     & = 
     D_{\mathrm{KL}} \left[q(\mathbf{z}) || p(\mathbf{z}|\mathcal{T}_i)\right].
    \end{aligned}
\end{equation}
This provides the final ELBO for the  variational task modulation: 
\begin{equation}
\label{eq:CVTI_elbo_2}
    \begin{aligned}
        q(\mathbf{z}, \hat{\mathbf{x}}^q, \hat{\mathbf{x}}^s | \mathcal{T}_i)
        & \geq \mathbb{E}_{q(\mathbf{z}, \hat{\mathbf{x}}^q, \hat{\mathbf{x}}^s)} \left[\log p(\hat{\mathbf{y}}| \hat{\mathbf{x}}^q, \hat{\mathbf{x}}^s)\right]  \\
        & -  D_{\mathrm{KL}} \left[q(\mathbf{z}) || p(\mathbf{z} | \mathcal{T}_i)\right] \\
    \end{aligned}
\end{equation} 
The overall computation graph of  variational task modulation is shown in Figure~\ref{fig:vi_graph}. 

Directly optimizing the above objective does not take into account the task information of all model layers, since it only focuses on the conditional latent scale and shift at a specific layer. Thus, we introduce hierarchical variational inference into the  variational task modulation by conditioning the posterior on both the known tasks and the conditional latent scale and shift from the previous layers.

\subsection{Hierarchical variational task modulation}
\label{sec:chvti}
We replace  variational distribution in eq.~(\ref{eq:first_kl}) with a new conditional distribution $q(\mathbf{z}^l, \hat{\mathbf{x}}^q, \hat{\mathbf{x}}^s |  \mathbf{z}^{l-1}, \mathcal{T}_j)$ that makes  latent scale and shift of  current $l$-th layer also dependent on the latent scale and shift from the upper $l{-}1$-th layers. 

\begin{figure} [t]
\centering
\includegraphics[width=0.9\linewidth]{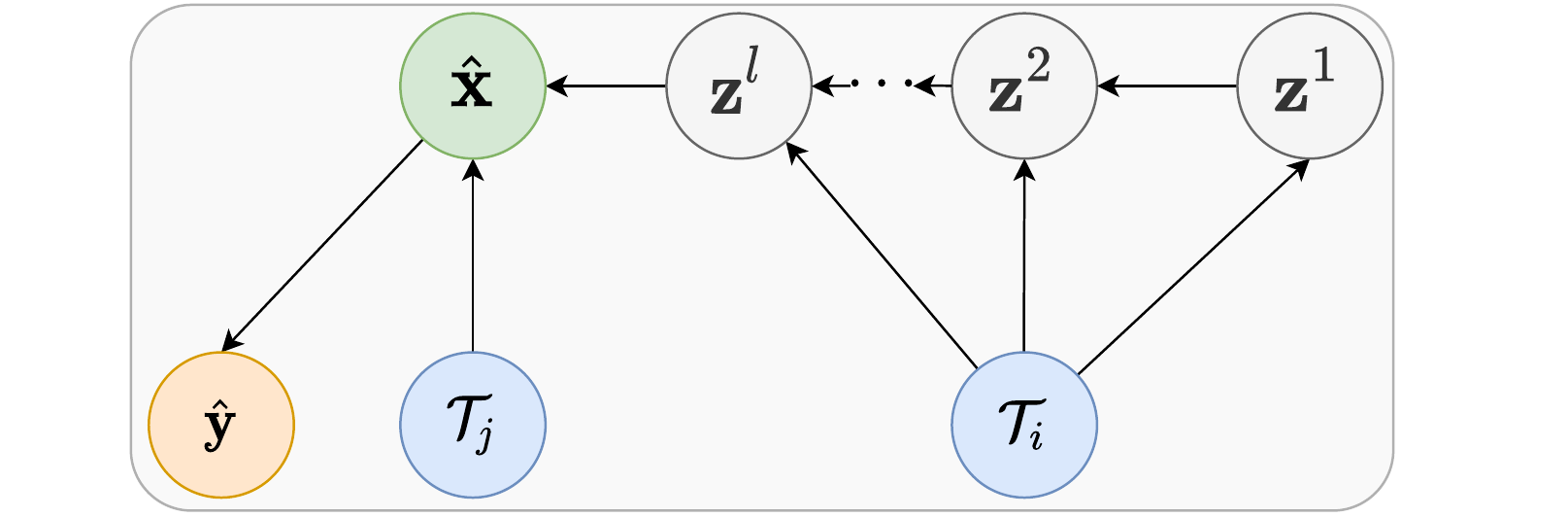}
\vspace{-4mm}
\caption{\textbf{Hierarchical variational task modulation.} $\mathbf{z}^l$ indicates the latent modulation parameters at the layer $l$. The latent transformation parameter $\mathbf{z}^l$ is depend on the task $\mathcal{T}_i$ and the upper $\mathbf{z}^{l-1}$.}
\vspace{-6mm}
	\label{fig:hvi_graph}
\end{figure}

The  hierarchical variational inference gives rise to a new ELBO, as follows:
\begin{equation}
\label{eq:CHVTI_elbo}
    \begin{aligned}
        q(\mathbf{z}, \hat{\mathbf{x}}^q, \hat{\mathbf{x}}^s | \mathcal{T}_i)
        & \geq \mathbb{E}_{q(\mathbf{z}^l, \hat{\mathbf{x}}^q, \hat{\mathbf{x}}^s | \mathbf{z}^{l-1})  } \left[\log p(\hat{\mathbf{y}}| \hat{\mathbf{x}}^q, \hat{\mathbf{x}}^s)\right]  \\
        & -  D_{\mathrm{KL}} \left[q(\mathbf{z}^l|\mathbf{z}^{l-1}) || p(\mathbf{z}^l |\mathbf{z}^{l-1},  \mathcal{T}_i)\right] \\
    \end{aligned}
\end{equation}
The graphical model of hierarchical variational task modulation is shown in Figure~\ref{fig:hvi_graph}.

In practice, the prior $p(\mathbf{z}^l |\mathbf{z}^{l-1},  \mathcal{T}_i)$ is implemented by an amortization network~\cite{kingma2013auto} that takes the concatenation of the average feature representations of samples in the support set from $\mathcal{T}_i$ and the upper layer latent scale and shift $\mathbf{z}^{l-1}$ 
and returns the mean and variance of the current layer latent scale and shift $\mathbf{z}^{l}$.  To enable back-propagation with the sampling operation during training, we adopt the reparametrization trick~\cite{rezende2014stochastic, kingma2013auto} as
$\mathbf{z} {=} \mathbf{z}_{\mu} + \mathbf{z}_{\sigma} \odot \boldsymbol\epsilon$, where $\mathbf\epsilon \sim \mathcal{N}(0, \mathrm{I} ).$ 
The hierarchical probabilistic scale and shift provide a more informative task representation than the deterministic meta task modulation and have the ability to capture different representation levels, thus modulating more diverse tasks for few-task meta-learning.  

In the meta-training stage, we use the known meta-training tasks $\mathcal{T}_i$  with our meta task modulation and its variational variants to generate the new tasks $\hat{\mathcal{T}}$ for the meta-training. 
To ensure that the original tasks are also trained together, we train the generated tasks together with the original tasks. Thus the loss function of our meta task modulation $\mathcal{L}_\mathrm{MTM}$ is as follows:
\begin{equation}
\small
\label{eq:CTI_final_loss}
    \begin{aligned}
        \mathcal{L}_\mathrm{MTM} = \frac{1}{T} \sum_{i}^{T}  \Big(  \sum_{(\hat{\mathcal{S}}_i, \hat{\mathcal{Q}}_i) \sim \hat{\mathcal{T}}_i}\mathcal{L}_{\mathrm{CE}} + \lambda \sum_{(\mathcal{S}_i, \mathcal{Q}_i) \sim \mathcal{T}_i} \mathcal{L}_{\mathrm{CE}} \Big).
    \end{aligned}
\end{equation}
The loss of variational task modulation $\mathcal{L}_\mathrm{VTM}$ is 
\begin{equation}
\small
\label{eq:CVTI_final_loss}
    \begin{aligned}
      \mathcal{L}_\mathrm{VTM} = &\frac{1}{T} \sum_{i, j}^{T}  \Big( \sum_{(\hat{\mathbf{x}}^q,  \hat{\mathbf{y}}) \in \hat{\mathcal{Q}}} - \mathbb{E}_{q(\mathbf{z}, \hat{\mathbf{x}}^q, \hat{\mathbf{x}}^s)} \left[\log p(\hat{\mathbf{y}}| \hat{\mathbf{x}}^q, \hat{\mathbf{x}}^s)\right] \\
        & +  \beta D_{\mathrm{KL}} \left[q(\mathbf{z}) || p(\mathbf{z} | \mathcal{T}_i)\right]
        \Big)   + \lambda \frac{1}{T} \sum_{i}^{T}  \sum_{(\mathcal{S}_i, \mathcal{Q}_i) \sim \mathcal{T}_i} \mathcal{L}_{\mathrm{CE}}.
    \end{aligned}
\end{equation}
And the loss of hierarchical variational task modulation can be written as
\begin{equation}
\small
\label{eq:CHVTI_final_loss}
    \begin{aligned}
        \mathcal{L}_\mathrm{HVTM} = &\frac{1}{T} \sum_{i, j}^{T}  \Big( \sum_{(\hat{\mathbf{x}}^q,  \hat{\mathbf{y}}) \in \hat{\mathcal{Q}}} -  \mathbb{E}_{q(\mathbf{z}^l, \hat{\mathbf{x}}^q, \hat{\mathbf{x}}^s | \mathbf{z}^{l-1})  } \left[\log p(\hat{\mathbf{y}}| \hat{\mathbf{x}}^q, \hat{\mathbf{x}}^s)\right]  \\
        & - \beta D_{\mathrm{KL}} \left[q(\mathbf{z}^l|\mathbf{z}^{l-1}) || p(\mathbf{z}^l |\mathbf{z}^{l-1},  \mathcal{T}_i)\right]
        \Big)   \\
        & + \lambda \frac{1}{T} \sum_{i}^{T}  \sum_{(\mathcal{S}_i, \mathcal{Q}_i) \sim \mathcal{T}_i} \mathcal{L}_{\mathrm{CE}},
    \end{aligned}
\end{equation}
where $\mathcal{L}_{\mathrm{CE}}$ is the cross-entropy loss,
\begin{equation}
\small
\label{eq:loss_ce}
    \begin{aligned}
        \mathcal{L}_{\mathrm{CE}} = \frac{1}{N_C N_{\mathcal{Q}}} \big[d(f_\phi(x^q), c_k) + \log \sum_{k'} \exp (-d(f_\phi(x^q), c_k))  \big],
    \end{aligned}
\end{equation}
$N_C$ and $N_{\mathcal{Q}}$ are the number of prototypes and query samples in each task,  and $\lambda > 0$ and  $\beta > 0$ are the regularization hyper-parameters.

In the meta-test stage, we directly input the support set $\mathcal{S}$ using the meta-trained feature extractor $f_{\phi}(\cdot)$ to obtain the prototype $c_k$  from the test task. Then we obtain the prediction of the query set $\mathbf{x}^q$ for performance evaluation based on eq.~(\ref{eqn:protonetProb}).


\section{Experiments}
\label{sec:experiments}

\subsection{Experimental setup}

\textbf{Datasets.} We conduct experiments on four few-task meta-learning challenges, \ie, miniImagenet, ISIC, DermNet and Tabular Murris~\cite{cao2020concept}. 
miniImagenet~\cite{vinyals16} is constructed from ImageNet~\cite{deng2009imagenet} and comprises a total of 100 different classes (each with 600 instances). All images are downsampled to 84 $\times$ 84. We follow \cite{yao2021meta} and reduce the number of tasks by limiting the number of meta-training classes to obtain miniImagenet-S, with 12 meta-training classes and 20 meta-test classes.
ISIC~\cite{milton2019automated} aims to classify dermoscopic images among nine different diagnostic categories. 10,015 images are available for training across 8 different categories. We select 4 categories as the meta-training classes.
DermNet is one of the largest open resources of images of skin diseases, with more than 23,000 images. Following \cite{yao2021meta}, we construct Dermnet-S, which selects 30 diseases as the meta-training classes.  
Tabular Murris considers cell type classification across organs and contains nearly 100,000 cells from 20 organs and tissues. Following \cite{yao2021meta}, we choose 57 base classes as the meta-training classes. For our ablation studies we report on miniImagenet-S, ISIC and Dermnet-S, for our comparison with the state-of-the-art, we also consider Tabular Murris. Sample images from all datasets are provided in the appendix.

\textbf{Implementation details.}
For miniImagenet-S, ISIC, DermNet-S and Tabular Murris, we follow \cite{yao2021meta} using a network containing four convolutional blocks and a classifier layer. Each block comprises a 32-filter 3 {$\times$} 3 convolution, a batch normalization layer, a ReLU nonlinearity, and a 2 {$\times$} 2 max pooling layer. We train a ProtoNet~\cite{snell2017prototypical} using Euclidean distance in the 1-shot and 5-shot scenarios with training episodes. Each image is re-scaled to the size of 84 {$\times$} 84 {$\times$} 3. For all experiments, we use an initial learning rate of $10^{-3}$ and an SGD optimizer with Adam~\cite{DiederikPKingma2014AdamAM}. The variational neural network is  parameterized by three feed-forward  multiple-layer perception networks  and a ReLU activation layer. The number of Monte Carlo samples is 20. 
The batch and query sizes of all datasets are set as 4 and 15. The total training iterations are 50,000. The average few-task meta-learning classiﬁcation accuracy (\%, top-1) is reported across all test images and tasks. Code  available at:~\url{https://github.com/lmsdss/MetaModulation}.

\subsection{Results}

\begin{table}[t]
\centering
\resizebox{0.9\linewidth}{!}{
\begin{tabular}{lcccccc}
\toprule
 & \multicolumn{2}{c}{\textbf{miniImagenet-S}} & \multicolumn{2}{c}{\textbf{ISIC}} & \multicolumn{2}{c}{\textbf{Dermnet-S}} \\
\cmidrule(){2-3} \cmidrule(lr){4-5} \cmidrule(lr){6-7} 
 & 1-shot & 5-shot & 1-shot & 5-shot & 1-shot & 5-shot\\
 \midrule
 Vanilla  & 36.26 &  50.72 & 58.56 & 66.25 & 44.21 & 60.33   \\
\rowcolor{Gray}
MTM & \textbf{42.44}  & \textbf{56.25} & \textbf{63.13 }& \textbf{74.23} &\textbf{49.46}  & \textbf{66.12}  \\
\bottomrule
\end{tabular}}
\vspace{-3mm}
\caption{\textbf{Benefit of meta task modulation} in (\%) on three few-task meta-learning challenges. Our meta task modulation (MTM) achieves better performance compared to a vanilla ProtoNet.}
\label{tab:ab_cti}
\vspace{-3mm}
\end{table}
\textbf{Benefit of meta task modulation.} 
To show the benefit of meta task modulation, we first compare our method with a vanilla Prototypical network~\cite{snell2017prototypical} on all tasks, without using task interpolation, in Table~\ref{tab:ab_cti}. Our model performs better under various shot conﬁgurations on all few-task meta-learning benchmarks. We then compare our model with the state-of-the-art MLTI~\cite{yao2021meta} in Table~\ref{tab:sota}, which interpolates the task distribution by Mixup~\cite{verma2019manifold}. 
Our meta task modulation also compares favorably to MLTI under various shot conﬁgurations. On ISIC, for example, we surpass MLTI by $2.71\%$ on the 5-way 5-shot setting.  This is because our model can learn how to modulate the base task features to better capture the task distribution instead of using linear interpolation as described in the ~\cite{yao2021meta}.

\begin{table}[t]
\centering
\resizebox{0.9\linewidth}{!}{
\begin{tabular}{lcccccc}
\toprule
& \multicolumn{5}{c}{\textbf{Network layer}} &  \\
\cmidrule(){2-6}
& $1^{\mathrm{st}}$ & $2^{\mathrm{nd}}$ & $3^{\mathrm{rd}}$ & $4^{\mathrm{th}}$ & random &  \textbf{All (\textbf{HVTM})} \\
 \midrule
\rowcolor{Gray}
5-way 1-shot & & & & & & \\
 MTM  & 41.30 &  41.32 & 41.31 & 39.47 & 39.98 & 42.44   \\
 VTM  & {41.25} & {42.05} & {41.63} & {39.97} & {40.91} & \textbf{43.21} \\
\midrule
\rowcolor{Gray}
5-way 5-shot & & & & & & \\
 MTM   &  54.21 & 54.30 & 54.13 & 52.62 & 53.32 & 56.25   \\
 VTM  & {54.47} & {55.82} & {54.36} & {52.80} & {54.43} & \textbf{57.26} \\
\bottomrule
\end{tabular}}
\vspace{-2mm}
\caption{\textbf{Benefit of  variational task modulation} for varying layers on miniImageNet-S. Variational task modulation (VTM) improves over any of the selected individual layers using MTM.}
\label{tab:ab_layer}
\vspace{-5mm}
\end{table}

\textbf{Benefit of variational task modulation.}
We investigate the benefit of variational task modulation by comparing it with deterministic meta task modulation. The results are reported on miniImageNet-S under various shots in Table~\ref{tab:ab_layer}.
$\{1^{\mathrm{st}}, 2^{\mathrm{nd}}, 3^{\mathrm{rd}}, 4^{\mathrm{th}}\}$, random and, all are  the selected determined layer, the randomly chosen one layer and  all  the layers to be modulated, respectively. 
The variational task modulation consistently outperforms the deterministic meta task modulation on any selected layers,  demonstrating the benefit  of probabilistic modeling. By using probabilistic task modulation, the base task can be modulated in a more informative way, allowing it to encompass a larger range of task distributions and ultimately improve performance on the meta-test task.

\begin{table}[t]
\centering
\resizebox{0.9\linewidth}{!}{
\begin{tabular}{lcccccc}
\toprule
& \multicolumn{2}{c}{\textbf{miniImagenet-S}} & \multicolumn{2}{c}{\textbf{ISIC}} & \multicolumn{2}{c}{\textbf{DermNet-S}} \\
\cmidrule(){2-3} \cmidrule(lr){4-5} \cmidrule(lr){6-7} 
 & 1-shot & 5-shot & 1-shot & 5-shot & 1-shot & 5-shot\\
\midrule
VTM  &  {42.05}  & {55.82} & {64.04}& {72.59} &{49.19}  & {64.62} \\
\rowcolor{Gray}
\textbf{HVTM}  & \textbf{43.21}  & \textbf{57.26} & \textbf{65.16}& \textbf{76.40} &\textbf{50.45}  & \textbf{67.05} \\
\bottomrule
\end{tabular}}
\vspace{-3mm}
\caption{\textbf{Hierarchical vs. flat  variational modulation.}
Hierarchical variational task modulation (HVTM) is more effective than flat variational task modulation (VTM) for few-task meta-learning.}
\label{tab:ab_vhcti}
\vspace{-5mm}
\end{table}

\textbf{Hierarchical vs. flat  variational task modulation.} We compare hierarchical modulation with flat  variational modulation, which only selects one layer to modulate. As shown in Table~\ref{tab:ab_vhcti}, the hierarchical variational modulation  improves the overall performance under both the 1-shot and 5-shot settings on all three benchmarks. 
The hierarchical structure is well-suited for increasing the density of the task distribution across different levels of features, which leads to better performance compared to flat variational modulation. This makes sense because the hierarchical structure allows for more informative transformations of the base task, enabling it to encompass a broader range of task distributions. Note that, we use hierarchical variational task modulation to compare the state-of-the-art methods in the subsequent experiments.

\begin{figure}[t!]
\centering
\includegraphics[width=0.8\linewidth]{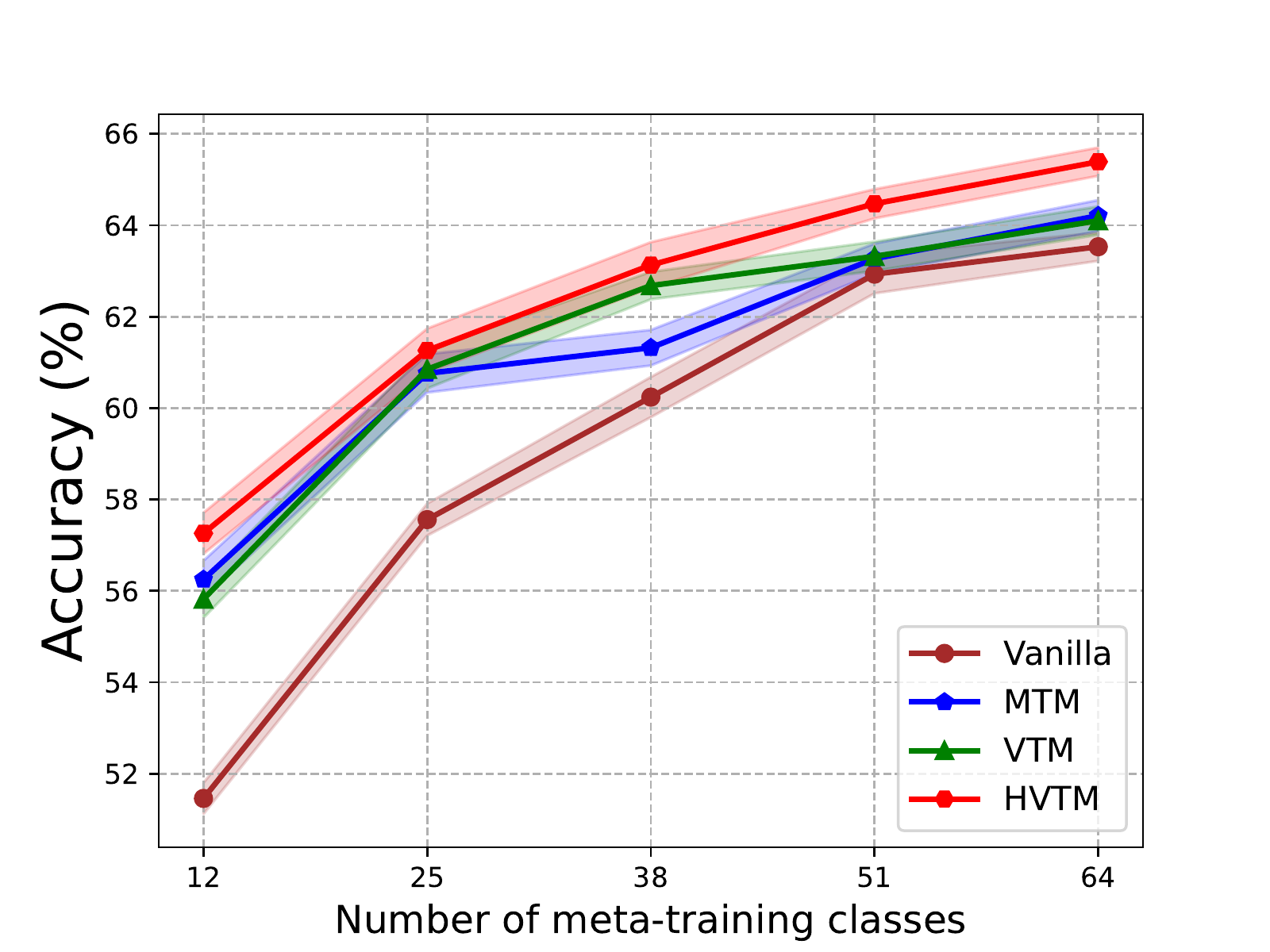}
\vspace{-3mm}
\caption{\textbf{Influence of the number of meta-training tasks} for 5-way 5-shot  on miniImageNet. All MetaModulation implementations improve over a vanilla prototype network, especially when fewer tasks are available for meta-learning. Where a vanilla network requires 64 tasks to reach 63.7\% accuracy, we need 40.}
\vspace{-3mm}
	\label{fig:task_number}
\end{figure}

\begin{table}[t]
\centering
\setlength{\tabcolsep}{1.2mm}{
\resizebox{0.85\linewidth}{!}{
\begin{tabular}{lcccc}
\toprule
  & \multicolumn{2}{c}{\textbf{mini $\rightarrow$ Dermnet}} & \multicolumn{2}{c}{\textbf{Dermnet $\rightarrow$ mini}}\\
\cmidrule(){2-3} \cmidrule(lr){4-5}
 & 1-shot & 5-shot & 1-shot & 5-shot\\\midrule
Vanilla & 33.12 & 50.13 & 28.11 & 40.35 \\
MLTI &  {35.46} & {51.79} &  {30.06} & {42.23} \\
ATA  & 35.83$\pm$0.58  & 51.65$\pm$0.6  & - & - \\ 
\rowcolor{Gray}
\textbf{\textit{This paper}}  & \textbf{37.15}$\pm$0.75 & \textbf{53.92} $\pm$ 1.01 &  \textbf{31.56} $\pm$ 0.68& \textbf{44.13} $\pm$ 0.92 \\
\bottomrule
\end{tabular}}
}
\vspace{-2mm}
\caption{\textbf{Cross-domain adaptation ability.} MetaModulation achieves better performance even in a challenging cross-domain adaptation setting compared to a vanilla prototype network and MLTI by~\citet{yao2021meta}.}
\vspace{-6mm}
\label{tab:cross_nls2}
\end{table}

\textbf{Influence of the number of meta-training tasks.}
In Figure~\ref{fig:task_number}, we analyze the effect of the number of available meta-training tasks on the performance of our model under a 5-shot setting on  miniImageNet-S. Naturally, our model's performance improves, as the number of meta-training classes increases. The number of meta-training tasks is important for making the model more generalizable through meta-learning. More interesting, our model's performance is considerably improved by using a learnable modulation that incorporates information from different levels of the task. Compared to the best result of a vanilla prototype network, 63.7\% for 64 meta-training classes, we can reduce the number of classes to 40 for the same accuracy.

\textbf{Cross-domain adaptation ability.} To further evaluate the effectiveness of our proposed method, we conducted additional tests to assess the performance of MetaModulation in cross-domain adaptation scenarios. We trained  MetaModulation on one source domain and then evaluated it on a different target domain. Specifically, we chose the miniImagenet-S and Dermnet-S domains. The results, as shown in Table~\ref{tab:cross_nls2}, indicate MetaModulation generalizes better even in this more challenging scenario. 

\begin{figure}[t!]
\centering
\includegraphics[width=0.8\linewidth]{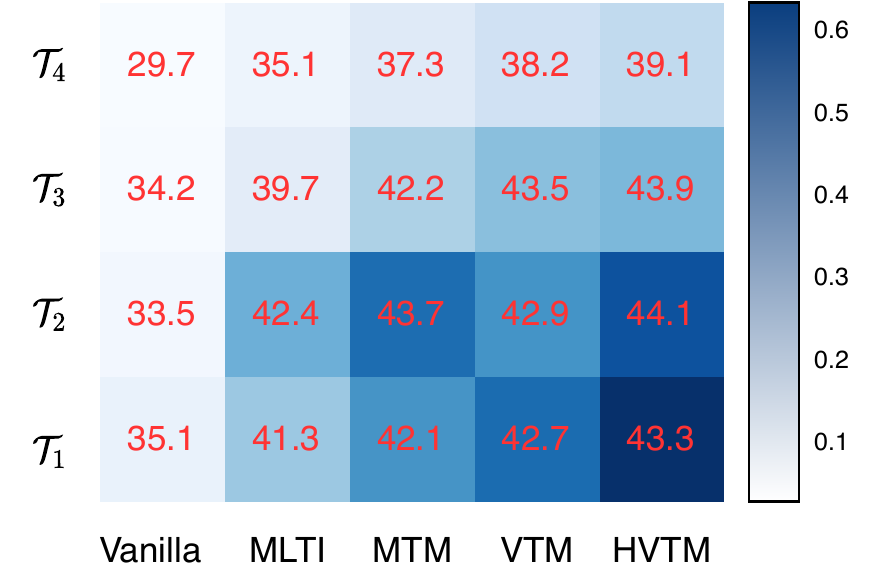}
\vspace{-3mm}
\caption{\textbf{Analysis of modulated tasks.}  Similarity of meta-training tasks to meta-test tasks for different methods, and the corresponding accuracy (\textcolor{red}{red} numbers) for the meta-test tasks. The tasks modulated by MetaModulatation have high similarity with the meta-test tasks, resulting in high accuracy.}
\vspace{-6mm}
	\label{fig:sim}
\end{figure}

\textbf{Analysis of modulated tasks.} To understand how our MetaModulation is able to improve performance, we plotted the similarity between the vanilla, interpolated and modulated tasks and the meta-test tasks in Figure~\ref{fig:sim}. Red numbers indicate the accuracy per model on each task. Specifically, we select 4 meta-test tasks and 300  meta-train tasks per model from the 1-shot miniImagenet-S setting to compute the task representation of each model.  
We then used instance pooling to obtain the representation of each task. Instance pooling involves combining a task's support and query sets and averaging the feature vectors of all instances to obtain a fixed-size prototype representation. This approach allows us to represent each task by a single vector that captures the essence of the task.
We calculated the similarity between meta-train and meta-test tasks using Euclidean distance. When using the vanilla prototype model~\cite{snell2017prototypical} directly, the similarity between meta-train and meta-test tasks is extremely low, indicating a significant difference in task distribution between meta-train and meta-test. This results in poor performance as seen in Figure~\ref{fig:sim} red numbers due to the distribution shift. However, the tasks modulated by our MetaModulation have a higher similarity with the meta-test tasks compared to the vanilla~\cite{snell2017prototypical} and MLTI~\cite{yao2021meta}, resulting in high accuracy. But, the similarity between the modulated tasks by our MetaModulation and $\mathcal{T}_4$ is also relatively low and performance is also poor. This may be because the task distribution of $\mathcal{T}_4$ is an outlier in the entire task distribution, making it hard to mimic this task during meta-training. Future work could investigate ways to mimic these outlier tasks in the meta-training tasks.


\begin{table*}[t]
\centering
\scalebox{0.75}{
\begin{tabular}{lcccccccc}
\toprule
 & \multicolumn{2}{c}{\textbf{miniImagenet-S}} & \multicolumn{2}{c}{\textbf{ISIC}} & \multicolumn{2}{c}{\textbf{Dermnet-S}}  & \multicolumn{2}{c}{\textbf{Tabular Murris}}\\
\cmidrule(){2-3} \cmidrule(lr){4-5} \cmidrule(lr){6-7}   \cmidrule(lr){8-9} 
& 1-shot & 5-shot & 1-shot & 5-shot & 1-shot & 5-shot  & 1-shot & 5-shot\\
\midrule
ProtoNet~\cite{snell2017prototypical}  & 36.26 &  50.72 & 58.56 & 66.25 & 44.21 & 60.33 & 80.03 &  89.20  \\
MAML~\cite{finn2017model} & 38.27 &  52.14 & 57.59 & 65.24 & 43.47 & 60.56 & 79.08 & 88.55  \\
Meta-Dropout~\cite{Lee2020Meta}  & 38.32 &  52.53 & 58.40 & 67.32 & 44.30 & 60.86 & 78.18 &  89.25  \\
TAML~\cite{jamal2019task} & 38.70 &  52.75 & 58.39 & 66.09 & 45.73 & 61.14 & 79.82 & 89.11  \\
MetaMix~\cite{yao2021improving} & 39.67 & 53.10 & 60.58 & 70.12 & 47.71 & 62.68 & 81.06 & 89.75\ \\
Meta-Maxup~\cite{yao2021improving} & 39.80 & 53.35 & 59.66 & 68.97 & 46.06 & 62.97 & 79.56 & 88.88  \\
Meta Interpolation~\cite{lee2022set} & {40.28} & {53.06} & - & - & - & -  & - & -\\
ATA~\cite{wang2023towards} & {40.62} & {54.59} & - & - & - & -  & - & -\\
MLTI~\cite{yao2021meta}  & {41.36} & {55.34} & {62.82} & {71.52} & {49.38} & {65.19} &81.89 & 90.12 \\
ATU~\cite{wu2022adversarial}  & {42.60} & {56.78} & {62.84} & {74.50} & {48.33} & {65.16} & 82.03 & 91.42 \\
\cmidrule{1-9}
\rowcolor{Gray}
This paper: \textit{\textbf{MetaModulation}} & \textbf{43.21}$\pm$0.73 & \textbf{57.26}$\pm$0.72 & \textbf{65.61}$\pm$1.09 & {\textbf{76.40}}$\pm$0.89 &\textbf{50.45}$\pm$0.84 & \textbf{67.05}$\pm$0.74 &\textbf{83.13}$\pm$0.89 & \textbf{91.23}$\pm$0.57 \\
\bottomrule 
\end{tabular}}
\caption{\textbf{Comparison with state-of-the-art.} All results, except for the MetaInterpolation~\cite{lee2022set}, are sourced from MLTI~\cite{yao2021meta}. MetaModulation is a consistent top performer for all settings and datasets. }
\vspace{-4mm}
\label{tab:sota}
\end{table*}

\textbf{Comparison with state-of-the-art.}
We evaluate MetaModulation on the four different datasets under 5-way 1-shot and 5-way 5-shot in Table~\ref{tab:sota}. Our model  achieves state-of-the-art performance on all four few-task meta-learning benchmarks under each setting.  On {miniImagenet-S, our model achieves 43.21\% under 1-shot, surpassing the second-best MLTI~\cite{yao2021meta}, by a margin of 1.85\%.   On ISIC~\cite{milton2019automated}, our method delivers 76.40\% for 5-shot, outperforming MLTI~\cite{yao2021meta} with 4.88\%. 
Even on the most challenging DermNet-S, which forms the largest dermatology dataset, our model delivers 50.45\% on the 5-way 1-shot setting. The consistent improvements on all benchmarks under various configurations confirm that our approach is effective for few-task meta-learning.

\section{Related work}
\label{sec:related}

\textbf{Few-task meta-learning.} 
In few-task meta-learning, the goal is to develop meta-learning algorithms that learn quickly and efficiently from a small number of examples with limited tasks in order to adapt to new tasks with minimal additional training.  
A common strategy for few-task meta-learning is task augmentation~\cite{yao2021improving, vu2021strata, murty2021dreca, zhou2021flipda, wang2021cross, wu2022adversarial, wang2023towards}, which adds additional tasks to the training data. One such approach is to generate additional tasks by perturbing the original tasks in some way~\cite{yao2021improving, murty2021dreca, zhou2021flipda, wu2022adversarial, wang2023towards}.
For example, MetaMix~\cite{yao2021improving} mixes support and query sets with Manifold Mixup~\cite{verma2019manifold} to construct a new query set. 
Another approach is to rely on unsupervised or self-supervised learning to generate additional tasks from the training data~\cite{vu2021strata, wang2021cross}. 
An alternative few-task meta-learning strategy is task interpolation~\cite{yao2021meta, lee2022set}, which trains a model to learn from a set of interpolated tasks.  
For example, MLTI~\cite{yao2021meta} performs Manifold Mixup on support and query sets from two tasks for task augmentation.  Set-based meta-interpolation~\cite{lee2022set} leverages expressive neural set functions~\cite{lee2019set} to interpolate a given set of tasks and trains the interpolating function using bilevel optimization so that the meta-learner trained with the augmented tasks generalizes to meta-validation tasks.  Both task augmentation and interpolation methods often randomly mix the features of two known tasks in a linear way without considering the features of other layers. This limits the diversity of the interpolated task and its potential benefit for increasing model generalizability. In contrast, we propose a learnable task modulation method that enables the model to learn a more diverse set of tasks by considering the features of each layer and allowing for a non-linear modulation between tasks.

\textbf{Conditional batch normalization.} 
Batch normalization~\cite{ioffe2015batch} is a crucial milestone in the development of deep neural networks. 
Conditional batch normalization (CBN)~\cite{de2017modulating} allows a neural network to learn different normalization parameters per class of input data.
Note the contrast to traditional batch normalization, which uses the same normalization parameters for all inputs to a network layer. By conditioning the normalization on additional information, such as the class labels of the training examples, CBN allows the network to adapt its normalization parameters to the specific class characteristics. 
Similarly, Perez \etal ~\cite{perez2018film} propose the feature-wise linear modulation layer for deep neural networks. 
In this paper, we take inspiration from conditional batch normalization and propose meta task modulation for few-task meta-learning, where the condition stems from the samples of a meta-training task. We use the conditional task as the condition, instead of data from another modality as in \cite{de2017modulating}, to predict the scale and shift parameters of the batch normalization for the base task.

\section{Conclusion}
\label{sec:conclusion}
In this paper, we addressed the issue of meta-learning algorithms requiring a large number of meta-training tasks which may not be readily available in real-world situations. We propose MetaModulation, which is to use a neural network to increase the density of the meta-training tasks by modulating batch normalization parameters during meta-training. Our MetaModulation consists of three different implementations. First is the meta task modulation, which modified parameters at various levels of the neural network to increase task diversity. Furthermore,  we  proposed a variational meta task modulation where the modulation parameters are treated as latent variables. We also introduced learning variational feature hierarchies by the variational meta task modulation. 
Our ablation studies showed the advantages of utilizing a learnable task modulation at different levels and the benefit of incorporating probabilistic variants in few-task meta-learning. 
Our MetaModulation and its variational variants consistently outperformed state-of-the-art few-task meta-learning methods on four few-task meta-learning benchmarks.

\section*{Acknowledgment}

This work is financially supported by the Inception Institute of Artificial Intelligence, the University of Amsterdam and the allowance 
Top consortia for Knowledge and Innovation (TKIs) from the Netherlands Ministry of Economic Affairs and Climate Policy, 
the National Key R\&D Program of China (2022YFC2302704), the Special Foundation of President of the Hefei Institutes of Physical Science (YZJJ2023QN06), and the Postdoctoral Researchers' Scientific Research Activities Funding of Anhui Province (2022B653).

\bibliography{egbib}
\bibliographystyle{icml2023}

\newpage
\appendix
\onecolumn
\section{Dataset.}
We apply our method to four few-task meta-learning image classiﬁcation benchmarks. Sample images from each dataset are provided in Figure~\ref{fig:dataset}.
\begin{figure*} [h]
\centering
\includegraphics[width=0.99\linewidth]{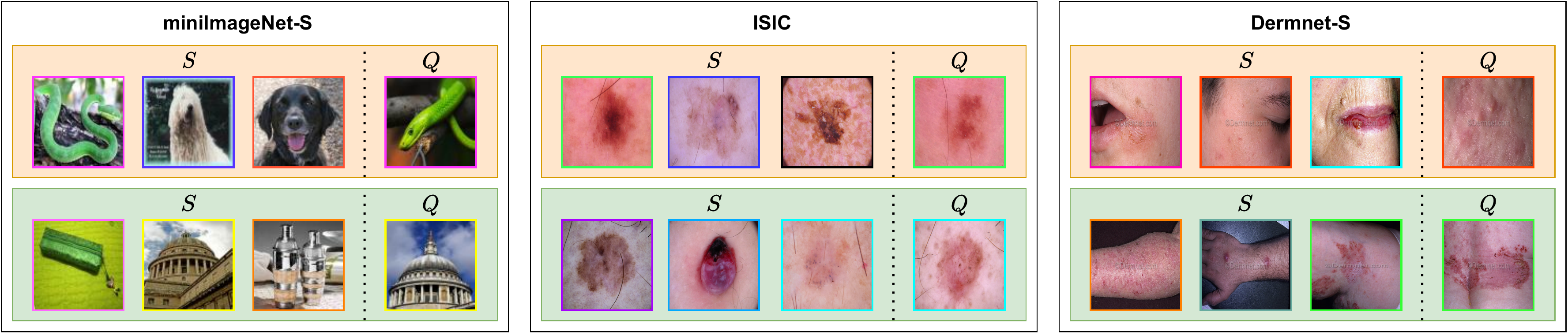}
\caption{Examples from each dataset. Orange and green boxes indicate the meta-training and meta-test tasks for each dataset.}
	\label{fig:dataset}
\end{figure*}

\section{Effect of the $\beta$.}
We test the impact of $\beta$ in (20) and (21). The value of $\beta$ control how much information in the base task will be modulated during the meta-training stage. The experimental results on the three datasets under both 1-shot and 5-shot setting are shown in Figure~\ref{fig:beta_1} and \ref{fig:beta_2}. We can see that the performance achieves the best when the values of $\beta$ are $0.01$. This means that in each modulate we need to keep the majority of base task. 

\begin{figure}[ht]
\centering
\begin{minipage}[b]{0.49\linewidth}
\includegraphics[width=\linewidth]{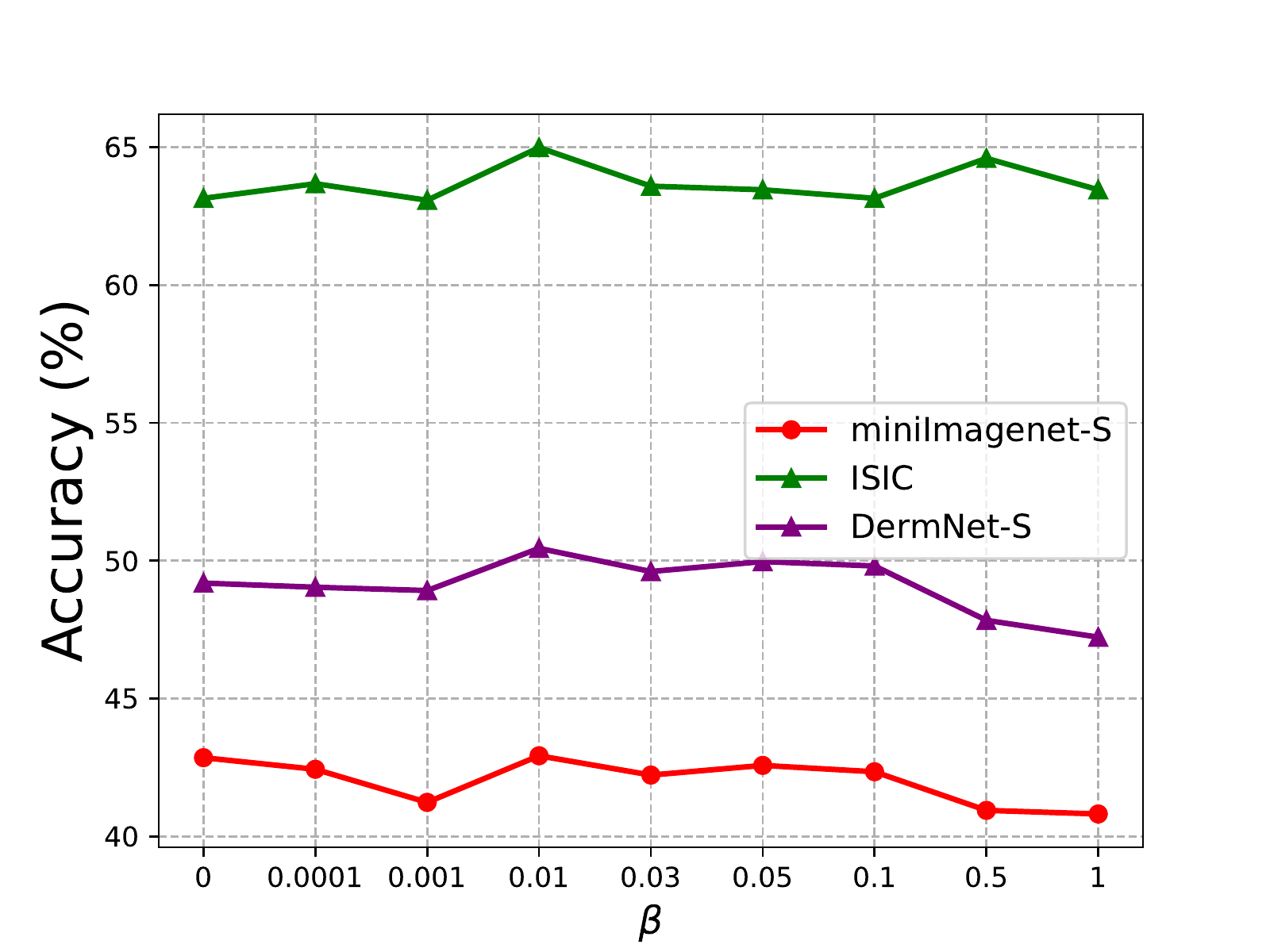}
\caption{Performance comparison by using various $\beta$ on the three few-task meta-learning dataset under 1-shot.}
\label{fig:beta_1}
\end{minipage}
\hfill
\begin{minipage}[b]{0.49\linewidth}
\includegraphics[width=\linewidth]{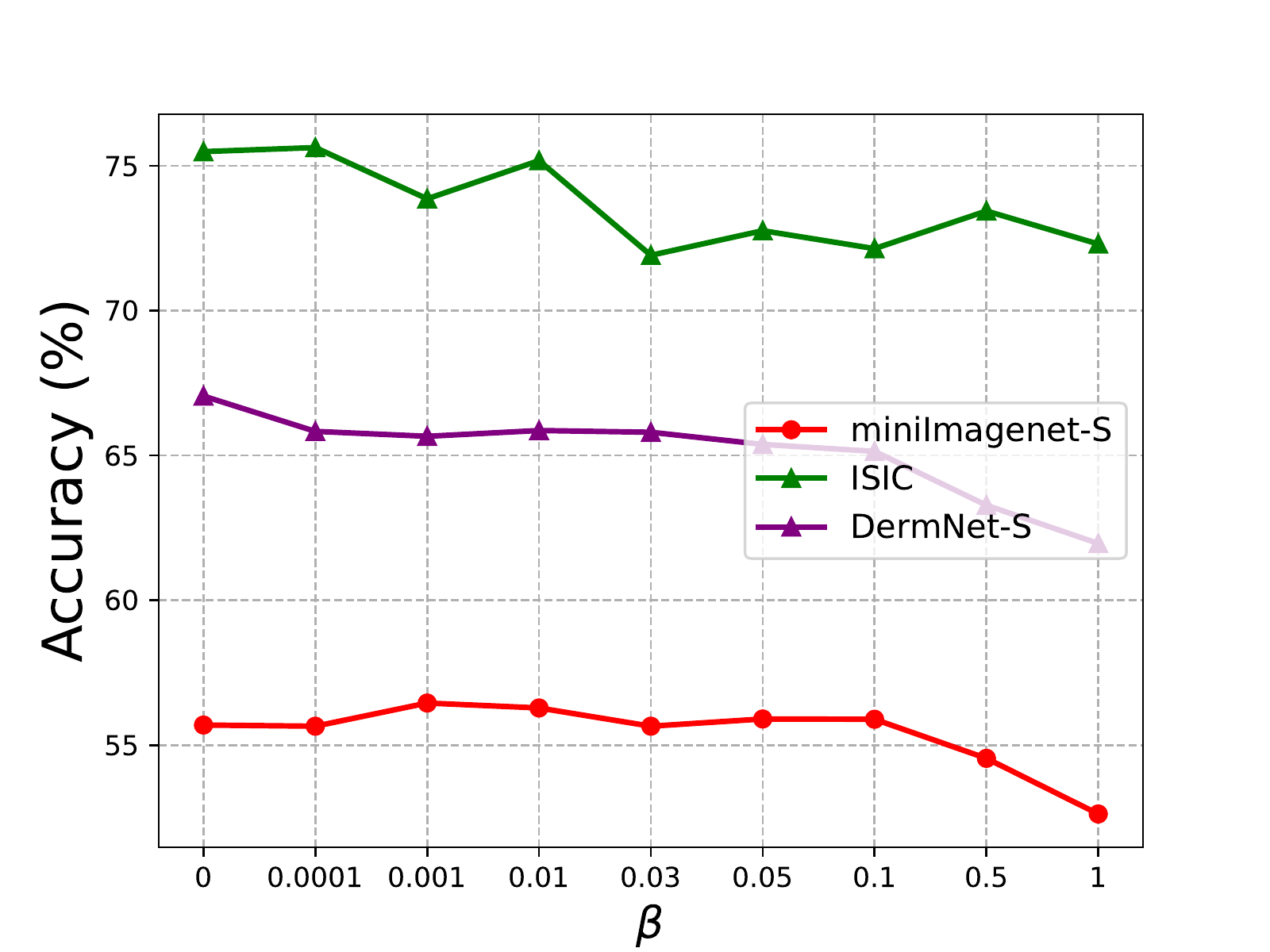}
\caption{Performance comparison by using various $\beta$ on the three few-task meta-learning dataset under 5-shot.}
\label{fig:beta_2}
\end{minipage}
\end{figure}

\section{Effect of the $\lambda$.}
We would like to emphasize that the hyper-parameters $\lambda$ (Eq. 19, 20, 21)  enable us to introduce constraints on new tasks, beyond just minimizing prediction loss.
By adjusting the value of  $\lambda$, we can control the trade-off between the prediction loss of the new tasks and the constraints imposed by the meta-training tasks. To clarify the impact of  $\lambda$, we performed an ablation on the HVTM (Eq. 21).
The results in Table~\ref{tab:lambda} show that when the original tasks have higher weight, the performance is worse. Additionally, we have conducted experiments to investigate the distribution differences between the meta-training and generated tasks. Specifically, in Table~\ref{tab:lambda}, we analyze the task representations of meta-training and generated tasks and show that they are similar, indicating that the generated tasks do indeed have a similar distribution as the meta-training tasks.

\begin{table}[t]
\centering
\scalebox{1}{
\begin{tabular}{lcccc}
\toprule
 & \multicolumn{2}{c}{\textbf{miniImagenet-S}} & \multicolumn{2}{c}{\textbf{ISIC}} \\
 \cmidrule(){2-3} \cmidrule(lr){4-5} 
& 1-shot & 5-shot & 1-shot & 5-shot \\
\midrule
\multirow{1}{*}{0.0001} & 41.97 & 55.23 & 65.25 & 76.23\\
\multirow{1}{*}{0.001} & 42.65 & 56.18 & 65.61 & 76.40 \\
\rowcolor{Gray}
\textbf{\textit{0.01}}  & \textbf{43.21} & \textbf{57.26} & \textbf{65.13} & \textbf{76.27} \\
\multirow{1}{*}{0.05} & 43.14 & 57.09 & 65.07 & 76.13 \\
\multirow{1}{*}{0.1} & 42.86 & 56.16 & 63.05 & 74.72 \\
\multirow{1}{*}{0} & 42.25 & 55.97 & 62.95 & 74.15 \\
\multirow{1}{*}{1} & 41.46 & 55.12 & 62.15 & 72.73 \\
\multirow{1}{*}{10} & 40.26 & 53.17 & 60.03 & 70.95 \\
\multirow{1}{*}{100} & 38.01 & 51.25 & 59.12 & 68.23 \\
\bottomrule
\end{tabular}}
\caption{\textbf{Ablation on the $\lambda$.}}
\label{tab:lambda}
\end{table}


\end{document}